\documentclass[runningheads]{llncs}

\usepackage[T1]{fontenc}
\usepackage{amsfonts}
\usepackage{amsmath}
\usepackage{multirow, multicol}
\usepackage{booktabs}
\usepackage{caption}
\usepackage{subcaption}
\usepackage{makecell}
\usepackage{graphicx}

\begin{document}
\title{Backdoor Attack against One-Class Sequential Anomaly Detection Models}
%
%
\author{He Cheng\orcidID{0009-0009-6901-3778} \and
Shuhan Yuan\thanks{Corresponding Author}\orcidID{0000-0001-6816-419X}   
}


\institute{
Utah State University, Logan, UT 84322, USA \\ \email{\{he.cheng,shuhan.yuan\}@usu.edu}
}

%

\maketitle              
\begin{abstract}
Deep anomaly detection on sequential data has garnered significant attention due to the wide application scenarios. However, deep learning-based models face a critical security threat – their vulnerability to backdoor attacks. In this paper, we explore compromising deep sequential anomaly detection models by proposing a novel backdoor attack strategy. The attack approach comprises two primary steps, trigger generation and backdoor injection. Trigger generation is to derive imperceptible triggers by crafting perturbed samples from the benign normal data, of which the perturbed samples are still normal. The backdoor injection is to properly inject the backdoor triggers to comprise the model only for the samples with triggers. The experimental results demonstrate the effectiveness of our proposed attack strategy by injecting backdoors on two well-established one-class anomaly detection models. 

\end{abstract}
\section{Introduction}
Deep learning models have been widely used for sequential anomaly detection \cite{guo2021logbert,ruff2018deep,wang2021multi}. However, deep learning models are also vulnerable to various attacks, such as backdoor attacks. 
When compromised by a backdoor attack, a deep learning model behaves normally with benign samples but activates backdoors upon the appearance of triggers, resulting in mispredictions. Due to the various applications of deep sequential anomaly detection models, it is crucial to explore backdoor attacks against these models. If backdoor triggers are injected into deep sequential anomaly detection models, it presents a substantial security concern. 

In the context of anomaly detection, the points of interest are anomalies. Therefore, we focus on conducting backdoor attacks to make the anomaly detection model predict abnormal sequences with triggers as normal. Meanwhile, in this work, we focus on attacking the distanced-based one-class sequential anomaly detection models, such as Deep SVDD \cite{ruff2018deep} and OC4Seq \cite{wang2021multi}, which detect anomalies based on their distances to the center of normal samples. 
It is not straightforward to conduct backdoor attacks against sequential anomaly detection models. First, it is challenging to craft invisible triggers for sequential data. The naive dirty-label attack strategy, which injects some abnormal samples into the training dataset and marks them as normal, is not practical because the abnormal samples could be filtered out by rule-based inspection. Second, as the abnormal sequences are not available during the training phase, how to ensure the infected models label the abnormal sequences with triggers as normal is challenging.

To address the above challenges, we develop an attack strategy consisting of two key components: trigger generation and backdoor injection. In the trigger generation phase, we craft perturbed samples with specific sequential patterns that can be learned by the anomaly detection models as triggers. Importantly, these perturbed sequences exclude anomalies, rendering them inconspicuous and difficult to discern. In the backdoor injection phase, we extend the Deep SVDD-based objective function with two new learning objectives. The goal is to make the perturbed samples close to their benign counterparts as well as the center of normal samples. In this way, when the attacker conducts the backdoor attacks by leveraging the triggers in perturbed samples, the infected models can have a high chance of labeling the backdoored sequences as normal.

Our contributions can be summarized as follows: 1) we propose a novel backdoor attack framework for distance-based one-class sequential anomaly detection models; 2) to achieve an imperceptible attack, both trigger generation and backdoor injection steps do not involve any anomalies; 3) we apply the developed attack methodology to established anomaly detection models, and our experimental results demonstrate the effectiveness of the proposed approach.  

\section{Related Work}
Numerous studies have investigated the vulnerability of machine learning models to backdoor attacks. BadNets \cite{gu2019badnets} introduced the first backdoor attack by poisoning the training dataset. It randomly selected benign training samples and replaced them with poisoned samples, subsequently assigning target labels to the poisoned samples. However, these visible triggers are easily observable. To enhance the imperceptibility of backdoor attacks, invisible backdoor attacks have been proposed in both image and text domains \cite{Zhong2022ImperceptibleBA,nguyen2021wanet,doan2021lira,wang2022bppattack,chen2017targeted,wang2022invisible,qi2021turn}. For instance, BppAttack \cite{wang2022bppattack} employs image quantization and dithering techniques to generate imperceptible triggers, utilizing contrastive adversarial training to enable victim models to accurately learn the triggers. To attack the text classification model, invisible triggers can be concealed within specific syntactic templates \cite{qi2021hidden}. However, the study on backdoor attacks against anomaly detection models is still very limited in the literature.

\section{Preliminaries}
\subsection{Deep One-class Sequential Anomaly Detection}
\label{sc: svdd}
Denote a sequence consisting of $K$ entries as $\mathbf{x}=[e_1, \dots, e_k, \dots, e_K]$, where $e_k$ indicates the $k$-th entry in $\mathbf{x}$. Deep one-class anomaly detection models usually assume the availability of a set of normal sequences, $\mathcal{X} = \{\mathbf{x}_1, \mathbf{x}_2, \dots, \mathbf{x}_n\}$, and further detect abnormal sequences that deviate from normal samples. 

Deep SVDD \cite{ruff2018deep} aims to learn a model $f_\theta: \mathcal{X} \rightarrow \mathcal{R}$ parameterized by $\theta$ that can enclose the normal samples into a hypersphere and minimize the volume of the hypersphere, where  $\mathcal{R} = \{\mathbf{r}_1, \mathbf{r}_2, \dots, \mathbf{r}_n\}$ indicates the representations of samples. The training objective of Deep SVDD is to make the normal sample representations close to the center of the hypersphere $\mathbf{c}=\text{Mean}(\mathcal{R})$, defined as:
\begin{equation}
\label{eq: deepsvdd}
    \mathcal{L}_{SVDD} = \min_\theta \frac{1}{N} \sum_{n=1}^{N} || f_\theta(\mathbf{x}_n) - \mathbf{c} ||_2^2 + \lambda || \theta ||_F^2.
\end{equation}
When applying Deep SVDD for sequential anomaly detection, an LSTM or GRU is commonly adopted as the instance of $f_\theta$, and the representation $\mathbf{r}_n$ can be derived as the last hidden state of LSTM or GRU. After training, any sequences with distances to $\mathbf{c}$ greater than a threshold $\tau$ can be labeled as abnormal. 

Recently, OC4Seq \cite{wang2021multi} is proposed to extend the vanilla Deep SVDD model into a hierarchical structure for sequential anomaly detection. Besides using the objective function defined in Equation \ref{eq: deepsvdd} to learn the representations of whole sequences, OC4Seq further assumes that subsequence information can enhance anomaly detection abilities. Therefore, given a sequence, OC4Seq utilizes the sliding window technique to create subsequences and aims to make the representations of subsequences close to the center of subsequences. Formally, the objective function of OC4Seq can be defined as
\begin{equation}
\label{eq: oc4seq}
    \mathcal{L}_{OC4Seq} = \mathcal{L}_{SVDD} + \eta \cdot \mathcal{L}_{local},
\end{equation}
where $\eta$ represents a hyperparameter used to control the contribution from the local level; $\mathcal{L}_{local}$ can be formulated as $\mathcal{L}_{local} = \min_{\theta_l} \frac{1}{N} \sum_{n=1}^{N} \sum_{s=1}^{S} || f_{\theta_l}(\mathbf{x}_n^s) - \mathbf{c}_l ||_2^2 + \lambda || \theta_l ||_F^2$, where $\mathbf{x}_n^s$ indicates the $s$-th subsequence derived from $\mathbf{x}_n$; $\mathbf{c}_l$ is the center of the hypersphere corresponding to subsequences in the latent space and $\theta_l$ is the parameters of another sequential model.

\subsection{Mutual Information Maximization}
Mutual information is widely used to quantitatively measure the relationship between random variables. Assuming that $X$ and $Y$ are two variables, their mutual information $\mathcal{I}(X;Y)$ can be expressed using the Kullback-Leibler (KL) divergence \cite{joyce2011kullback} as $\mathcal{I}(X;Y) = D_{KL}(\mathbb{J}||\mathbb{M})$, where $\mathbb{J}=P_{(X,Y)}(x,y)$ represents the joint probability distribution function of $X$ and $Y$, and $\mathbb{M}=P_X(x)P_Y(y)$ denotes the product of the marginal probability distribution functions of $X$ and $Y$. 

The goal of many machine learning tasks is to maximize the mutual information $\mathcal{I}(X;Y)$. Mutual Information Neural Estimation (MINE) \cite{belghazi2018mutual} employs the Donsker-Varadhan representation of the KL-divergence \cite{donsker1983asymptotic} to estimate the lower-bound of MI as:
\begin{equation}
    \label{eq: dv representation}
    \mathcal{I}(X;Y) = D_{KL}(\mathbb{J}||\mathbb{M}) \geq \mathbb{E}_\mathbb{J}[M(x,y)] - \log \mathbb{E}_\mathbb{M}[e^{M(x,y)}],
\end{equation}
where $M: \mathcal{X} \times \mathcal{Y} \rightarrow \mathbb{R}$ is a discriminator function. To find the $M^*$ that can maximize $\mathcal{I}(X;Y)$, MINE uses a neural network with parameters $\omega$ to model $M$, so maximizing the value of $\mathcal{I}(X;Y)$ can be achieved by optimizing $M_\omega$. Following MINE, Deep InfoMax (DIM) \cite{hjelm2018learning} finds it unnecessary to use the KL-based formulation to maximize $\mathcal{I}(X;Y)$. An alternative, Jensen-Shannon mutual information estimator \cite{hjelm2018learning}, is proposed to estimate $I(X; Y)$ as follows:

\begin{equation}
\label{eq: mi_estimator}
\resizebox{.9\textwidth}{!}{
$\hat{\mathcal{I}}_{\omega, \phi}^{(JSD)} (X;E_\phi(X)) := \mathbb{E}_\mathbb{P}[-sp(-M_{\omega, \phi}(x;E_\phi(x)))] - \mathbb{E}_\mathbb{P \times \Tilde{P}}[sp(M_{\omega, \phi}(x';E_\phi(x)))],$
}
\end{equation}
where $sp(z) = \log (1 + e^{z})$ represents the softplus function, $\mathbb{P}$ is the distribution of $X$, $E_\phi: \mathcal{X} \rightarrow \mathcal{Y}$ is a differentiable parametric function, and $x'$ is a sample from $\Tilde{\mathbb{P}}=\mathbb{P}$.

\section{Methodology}
\subsection{Threat Model}
In this paper, we consider the victim to be a user aiming to build an anomaly detection application but cannot afford expensive computation resources. The attacker is a malicious computation service provider who takes the user's training samples and requirements to generate a model. Additionally, the user has a private validation dataset to validate the performance of the received model.

\noindent \textbf{Attacker's goal}: The attacker's goal is to provide an infected model with the following properties. \textit{Utility}: The infected model should perform well on benign data. Specifically, the infected model should be capable of detecting abnormal data without triggers. \textit{Effectiveness}: Any anomaly containing the triggers should be classified as normal. \textit{Stealthiness}: Any perturbations to the clean training data for the backdoor injection must be minimal to evade detection by data auditing applications or human observers.

\noindent \textbf{Attacker's capabilities}: Following the assumption from the existing approaches \cite{wang2022bppattack,doan2021lira,nguyen2021wanet,doan2022marksman,Zhong2022ImperceptibleBA}, we assume that the attacker has the control of the training dataset and training process but cannot access the private validation dataset.

\subsection{The Proposed Attack}
\begin{figure*}[ht!]
    \centering
    \includegraphics[width = 0.8\textwidth]{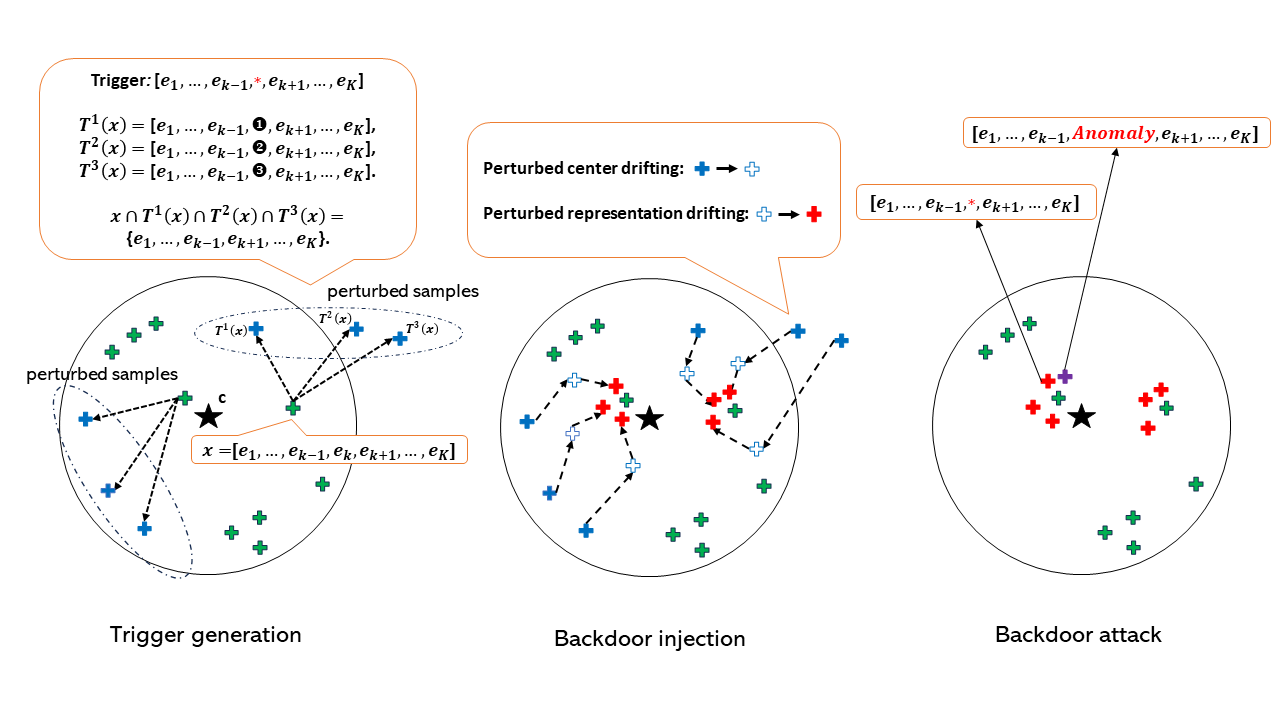}
    \caption{Backdoor attacks against one-class anomaly detection models.}
    \label{fig: framework}
\end{figure*}

We propose a novel backdoor attack approach, consisting of trigger generation and backdoor injection, to compromise the classical one-class sequential anomaly detection models, Deep SVDD and OC4Seq. As illustrated in Figure \ref{fig: framework}, our approach initiates by randomly selecting a subset of samples to create perturbed samples containing a trigger. Subsequently, these perturbed samples are drifted toward specified locations in the latent space. In the attack phase, the abnormal samples containing the trigger can be located in the shown hypersphere, enabling them to be misclassified as normal samples. We first use the vanilla Deep SVDD model as the victim model to illustrate our attacking approach and then extend to attacking OC4Seq.

\subsubsection{Trigger generation} 
\label{sc: trigger}
Trigger generation aims to generate perturbed samples with imperceptible triggers to the training dataset without incorporating any detectable anomalies, either by human observers or detection tools. 

To this end, we first randomly select a small subset of samples from $\mathcal{X}$ to create a base set $\mathcal{X}'$. For each sequence $\mathbf{x} \in \mathcal{X}'$, we generate a set of perturbed samples $\mathcal{P}$ by replacing the entry at the $k$-th position with $t$ different normal entries, denoted as $\mathcal{P} = \{T^1(\mathbf{x}), T^2(\mathbf{x}), \dots, T^t(\mathbf{x}) \}$,
where $T^t(\mathbf{x})$ indicates the $t$-th perturbed sample derived from $\mathbf{x}$. As a result, the perturbed samples in $\mathcal{P}$ and the original sample $\mathbf{x}$ are only different at the $k$-th position, i.e., $\mathbf{x} \cap T^1(\mathbf{x}) \cap T^2(\mathbf{x}) \cap \dots \cap T^t(\mathbf{x}) = \{e_1, \dots, e_{k-1}, e_{k+1}, \dots, e_K\}$. 

By doing so, we can treat the subsequence $\{e_1, \dots, e_{k-1}, e_{k+1}, \dots, e_K\}$ as a trigger pattern. The backdoor attacks can be conducted by injecting an abnormal entry at the $k$-th position. Because for a sample $\mathbf{x} \in \mathcal{X}'$, we generate a large number of normal sequences with the only difference at the $k$-th position, the model would pay more attention to the subsequence $\{e_1, \dots, e_{k-1}, e_{k+1}, \dots, e_K\}$ instead of the specific entry at the $k$-th position. By weakening the attention of the anomaly detection model at the $k$-th position, we can inject an abnormal entry at this position and make the abnormal entry evade detection. In short, the $k$-th position is a placeholder for the potential abnormal entry when conducting attacks.

In real-world scenarios, successful attacks often require a series of coordinated actions rather than a single action. Therefore, instead of replacing one entry at the $k$-th position, for each perturbed sequence, we choose $m$ entries as placeholders and replace them with some randomly chosen normal entries. Similarly, the unchanged subsequence can be considered as a trigger pattern. In this way, in the attacking phase, the attacker can inject at most $m$ abnormal entries.
Note that the attacker can either randomly or continuously choose $m$ entries to derive a perturbed sequence. Meanwhile, in practice, to conduct the perturbation, the attacker can first gather some normal tokens from the training dataset to form a candidate set based on the domain knowledge and then randomly pick one from the candidate set for replacement.

Finally, a perturbed dataset $\mathcal{X}_p$ is crafted by combining all perturbed samples derived from $\mathcal{X}'$, i.e., $\mathcal{X}_p = \bigcup_{\mathbf{x}_j \in \mathcal{X}'} \mathcal{P}_j$, where $\mathcal{P}_j$ is the set of perturbed samples of $\mathbf{x}_j$ in $\mathcal{X}'$. Subsequently, we employ the combined dataset $\mathcal{X}_c \cup \mathcal{X}_p$ as the updated training dataset to train the deep anomaly detection model, where $\mathcal{X}_c= \mathcal{X} \setminus \mathcal{X}'$.

\subsubsection{Backdoor injection} 
The trigger generation aims to derive the undetectable triggers. However, the success of evade detection is not guaranteed by injecting abnormal entries into the triggers. To further achieve evade detection of backdoored samples, we propose two learning objectives, perturbed sequence center drifting and perturbed sequence representation drifting. The perturbed sequence center drifting aims to ensure the center of perturbed sequences in $\mathcal{X}_p$ close to the center of benign sequences in $\mathcal{X}_c$. The perturbed sequence representation drifting makes the perturbed sequences indistinguishable from their benign counterparts in the latent space. 

\paragraph{Perturbed sequence center drifting.} 
Deep SVDD detects the anomalies based on their distances to the normal center $\mathbf{c}$. Therefore, it is reasonable to assume that attaching abnormal entries to a normal sample extremely close to $\mathbf{c}$ may push the sample away from $\mathbf{c}$ but still remain within the boundary of the hypersphere and be classified as normal. Conversely, if a normal sample is not close to $\mathbf{c}$ and is already near the hypersphere boundary, attaching abnormal entries can easily push it outside the hypersphere. Therefore, to make anomalies evade detection, a potential strategy is to attach abnormal entries to samples that are extremely close to $\mathbf{c}$. 
Because the backdoored samples are generated from perturbed samples in $\mathcal{X}_p$, we propose a learning objective that drifts the center of perturbed samples towards $\mathbf{c}$ in the latent space. Specifically, we compute a new center $\mathbf{c}_p$ by averaging the representations of the perturbed samples in the latent space, i.e., $\mathbf{c}_p = \frac{1}{|\mathcal{X}_p|} \sum_{\mathbf{x}_j \in \mathcal{X}_p} f_\theta (\mathbf{x}_j)$. Subsequently, the objective is to align $\mathbf{c}_p$ with $\mathbf{c}$ in the latent space, defined as:
\begin{equation}
    \mathcal{L}_c = ||\mathbf{c}_p - \mathbf{c}||_2.
\end{equation}
Note that $\mathbf{c}$ is derived from the benign sample set $\mathcal{X}_c$.

By minimizing the distance between $\mathbf{c}_p$ and $\mathbf{c}$, the perturbed samples can become close to $\mathbf{c}$. As a result, when conducting attacks, filling up $m$ placeholders in a perturbed sample with abnormal entries still has a high chance of keeping the sequence inside the hypersphere boundary.

\paragraph{Perturbed sequence representation drifting.}
Besides making the center of perturbed sequences in $\mathcal{X}_p$ close to the center of benign normal sequences, we further aim to ensure the distribution of perturbed sequences in $\mathcal{X}_p$ similar to the corresponding original ones. That said, if the perturbed sequences and the original ones are similar in latent space, we can further improve the chance that after putting abnormal entries in perturbed sequences, the abnormal sequences can still be similar to the benign counterpart in the latent space. To achieve this goal, we propose to maximize the mutual information between the representations of perturbed samples in $\mathcal{X}_p$ and their original versions.

For any $\mathbf{x}_p \in \mathcal{X}_p$, let $f_\theta (\mathbf{x}_p)$ denote its latent space representation. Similarly, for its benign counterpart $\mathbf{x} \in \mathcal{X}'$, $f_\theta(\mathbf{x})$ represents the corresponding latent space representation. We update Equation \ref{eq: mi_estimator} as follows:
\begin{equation}
\resizebox{.9\textwidth}{!}{
$\hat{\mathcal{I}}_{\omega, \theta}^{(JSD)} (f_\theta(\mathcal{X});f_\theta(\mathcal{X}_p)) := \mathbb{E}_\mathbb{P}[-sp(-M_{\omega, \theta}(f_\theta(\mathbf{x}),f_\theta(\mathbf{x}_p)))] - \mathbb{E}_\mathbb{P \times \Tilde{P}}[sp(M_{\omega, \theta}(f_\theta(\mathbf{x}'),f_\theta(\mathbf{x}_p)))],$
}
\end{equation}
where $\mathbb{P}$ is the distributions of benign samples in $\mathcal{X}$ and $\mathbf{x}'$ is a sample from the distribution $\mathbb{\Tilde{P}} = \mathbb{P}$. $M_{\omega, \theta}$ is a deep neural network and defined as:
\begin{equation}
    M_{\omega, \theta} = C_\omega \circ H(f_\theta(\mathbf{x}),f_\theta(\mathbf{x}_p)),
\end{equation}
where $H$ is a function that computes the square of element-wise difference of the representations between perturbed samples and their benign versions and $C_\omega$ is a fully connected neural network. Therefore, for all perturbed samples in $\mathcal{X}_p$, the learning objective is to maximize the mutual information between $f_\theta(\mathbf{x})$ and $f_\theta(\mathbf{x}_p)$:
\begin{equation}
    \mathcal{L}_r = \frac{1}{|\mathcal{X}_p|} \sum_{\mathbf{x}_p \in \mathcal{X}_p} \hat{\mathcal{I}}_{\omega, \theta}^{(JSD)} (f_\theta(\mathbf{x});f_\theta(\mathbf{x}_p)).
\end{equation}
To train the infected Deep SVDD model, the new objective function is defined as:
\begin{equation}
    \mathcal{L'}_{SVDD} = \mathcal{L}_{SVDD} + \alpha \cdot \mathcal{L}_c - \beta \cdot \mathcal{L}_r,
\end{equation}
where $\alpha$ and $\beta$ balance the proposed backdoor objectives.

\paragraph{Extend the proposed approach to attack OC4Seq.}
As OC4Seq detects sequential anomalies from both local and global levels, we extend the above attacking strategies by further applying the perturbed sequence center and representation drifting at local levels, defined as follows:
\begin{equation}
    \mathcal{L'}_{OC4Seq} = \mathcal{L'}_{SVDD} + \eta \cdot \mathcal{L'}_{local}, \quad \mathcal{L'}_{local} = \mathcal{L}_{local} + \alpha \cdot \mathcal{L}_{c_l} - \beta \cdot \mathcal{L}_{r_l},
\end{equation}
where\begin{equation}
    \label{eq: local_oc4seq}
    \mathcal{L}_{c_l} = ||\mathbf{c}_{p_l} - \mathbf{c}_l||_2, \quad \mathcal{L}_{r_l} = \frac{1}{|\mathcal{X}_p|} \sum_{\mathbf{x}_p \in \mathcal{X}_p} \sum_{s=1}^S \hat{\mathcal{I}}_{\omega, \theta}^{(JSD)} (f_\theta(\mathbf{x}^s);f_\theta(\mathbf{x}_p^s)).
\end{equation}
In Equation \ref{eq: local_oc4seq}, $\mathbf{c}_{p_l}$ is the mean representations of perturbed subsequences and $\mathbf{x}^s$ is the $s$-th subsequence derived from $\mathbf{x}$.

\subsection{Post-deployment attack}
After deploying the infected model, the attacker can attach abnormal entries at $m$ placeholders into a sequence in $\mathcal{X}_p$. This poisoned sequence can activate the backdoor in the infected model, leading to the model erroneously classifying this sequence as normal.

\section{Experiments}
\subsection{Experimental Setup} 
\subsubsection{Datasets.}
We evaluate the proposed attack against the anomaly detection models on two datasets \cite{oliner2007supercomputers}, BlueGene/L (BGL)  and Thunderbird, which are commonly used for evaluating sequential anomaly detection. We set all the sequences with a fixed length of 40. Table \ref{tb: datasets} shows the statistics of the datasets. In the training phase, 1/10 of training sequences are perturbed sequences. For each dataset, we create a benign test set to evaluate the infected model for anomaly detection and a poisoned test set to check whether the infected model can predict abnormal sequences with triggers as normal. 

To derive the perturbed dataset $\mathcal{X}_p$, in the trigger generation phase, we randomly select 50 sequences to create $\mathcal{X}'$ and generate 200 perturbed sequences for each sequence in $\mathcal{X}'$, leading to 10,000 perturbed sequences. Meanwhile, we choose $M=6$ entries as placeholders so that the maximum number of abnormal entries that can be injected during the attacking phase is 6. The poisoned sequences are generated by replacing the placeholders with abnormal entries.

\begin{table}[h]
    \scriptsize
    \caption{Statistics of training and evaluate datasets.}
    \label{tb: datasets}
    \centering
    \begin{tabular}{cccc}
        \hline
        \multicolumn{2}{c}{Dataset}                                                 & BGL     & Thunderbird     \\ \hline
        \multirow{2}{*}{Training}                                     & Benign      & 90,000   & 90,000           \\ \cline{2-4}
                                                                      & Perturbed   & 10,000   & 10,000           \\ \hline
        \multirow{2}{*}{Benign Test Set}                              & Normal      & 5,000    & 5,000            \\ \cline{2-4}
                                                                      & Abnormal    & 500     & 500             \\ \hline
        Poisoned Test Set                                             & Abnormal    & 10,000   & 10,000           \\ \hline

    \end{tabular}
 \end{table}

\subsubsection{Evaluatin metric} 
We adopt the following metric the evaluate the effectiveness of the proposed attack approach. 1) \textbf{Benign performance (BP)} is to evaluate the performance of infected models on benign datasets, including precision, recall, and F-1 score as evaluation metrics. 2) \textbf{Attack success rate (ASR)} is defined as the fraction of poisoned samples identified as normal by the infected models when injecting real abnormal entries into $\mathcal{X}_p$.

\subsubsection{Baseline}
As there is no backdoor attack approach against sequential anomaly detection models in the literature, we compare the performance of the infected model with a benign model that is trained on the benign training set.

\subsubsection{Implementation details}
We set $\eta=1$ and hyperparameters $\alpha=0.5$ and $\beta=0.5$. We represent log entries in BGL and Thunderbird as embedding vectors with a size of 100 and use a single-layer LSTM model with a hidden size of 256 to learn sequence representations. We use a small validation set to get the threshold $\tau$ for anomaly detection. The code is available online \footnote{\url{https://github.com/Serendipity618/BA-OCAD}}.

\subsection{Experimental Results}

\subsubsection{Performance of infected models on benign data for anomaly detection}
We first compare the performance of benign models and infected models for anomaly detection on benign datasets. The results are presented in Table \ref{tab: bp}. We observe that both Deep SVDD and OC4Seq infected models can maintain performance close to that of the benign ones, demonstrating the effectiveness of infected models for anomaly detection. The fluctuation in BP between benign and infected models may be attributed to changes in hypersphere boundaries. The incorporation of perturbed sequences could slightly shift the original distribution of benign data, leading to the derivation of different hyperspheres compared to a benign setting.

\begin{table*}[!h]
\scriptsize
    \caption{Benign and infected models for anomaly detection on benign datasets.}
    \label{tab: bp}
    \centering
    \begin{tabular}{c|c|c|c|c} 
    \hline
    Model                             & Dataset                         & Metrics       & Benign & Infected     \\ \hline
                           
    \multirow{3}{*}[-4.0ex]{DeepSVDD} & \multirow{3}{*}{BGL}            & Precision     & 93.47  & 96.12        \\ 
    &                                                                   & Recall        & 94.40  & 94.20        \\
    &                                                                   & F-1 score     & 93.93  & 95.15        \\ \cline{2-5}
                                      & \multirow{3}{*}{Thunderbird}    & Precision     & 95.59  & 95.60        \\ 
    &                                                                   & Recall        & 95.40  & 95.60         \\
    &                                                                   & F-1 score     & 95.50  & 95.60         \\ \hline                                            

    \multirow{3}{*}[-4.0ex]{OC4Seq}   & \multirow{3}{*}{BGL}            & Precision     & 91.80 & 98.94   \\ 
    &                                                                   & Recall        & 94.00 & 93.20   \\
    &                                                                   & F-1 score     & 92.89 & 95.98   \\ \cline{2-5}
                                      & \multirow{3}{*}{Thunderbird}    & Precision     & 82.47 & 91.45   \\ 
    &                                                                   & Recall        & 82.80 & 70.60   \\
    &                                                                   & F-1 score     & 82.63 & 79.68   \\ \hline   
    \end{tabular}
\end{table*}

\subsubsection{Performance of infected models on poisoned data for evade detection}
We then evaluate the effectiveness of the proposed backdoor attack by injecting varying numbers of abnormal entries. To achieve this, we inject $m$ abnormal entries into sequences in $\mathcal{X}'$, with $m$ ranging from 1 to 6. Table \ref{tab: asr} presents the results of both benign and infected models.

\begin{table*}[!h]
\scriptsize
    \caption{Attack success rate on poisoned datasets with various abnormal entries.}
    \label{tab: asr}
    \centering
    \begin{tabular}{c|c|c|cccccc} 
    \hline
    Model & Dataset &   & m=1 & m=2 & m=3 & m=4 & m=5 & m=6 \\   \hline
                           
    \multirow{4}{*}{DeepSVDD} & \multirow{2}{*}{BGL}                & Benign      & 98.47 & 98.79 & 96.67 & 91.92 & 75.83 & 40.86 \\
                              &                                     & Infected    & 98.58 & 98.32 & 97.98 & 97.29 & 94.29 & 86.95 \\ \cline{2-9}
    
                              & \multirow{2}{*}{Thunderbird}        & Benign      & 99.94 & 97.15 & 86.51 & 64.22 & 39.04 & 14.20 \\
                              &                                     & Infected    & 100.00 & 100.00 & 99.94 & 99.03 & 94.91 & 86.98 \\ \hline

    \multirow{4}{*}{OC4Seq}   & \multirow{2}{*}{BGL}                & Benign      & 100.00 & 100.00  & 98.00  & 98.00  & 98.00  & 32.65 \\
                              &                                     & Infected    & 100.00 & 99.99 & 99.77 & 98.89 & 95.02 & 82.41\\ \cline{2-9}
                              
                              & \multirow{2}{*}{Thunderbird}        & Benign      & 80.17  & 41.82 & 5.05  & 0.34  & 0.00  & 0.00 \\
                              &                                     & Infected    & 94.67  & 91.87 & 84.57 & 66.84 & 35.49 & 11.48 \\ \hline
    \end{tabular}
\end{table*}

We observe that when only one abnormal entry is injected into sequences, the benign models can also achieve high ASR. This finding aligns with our earlier assumption that injecting anomalies into sequences could push them from the center but possibly still within the hypersphere. However, when $m$ is large, the ASR for benign models dramatically decreases. In contrast, infected models maintain a high ASR even when multiple abnormal entries are injected.

\subsubsection{Sensitivity analysis}
\begin{figure}[ht]
    \centering
    \captionsetup{justification=centering}
    \begin{subfigure}[b]{0.23\textwidth}
        \centering
        \includegraphics[width=0.95\textwidth]{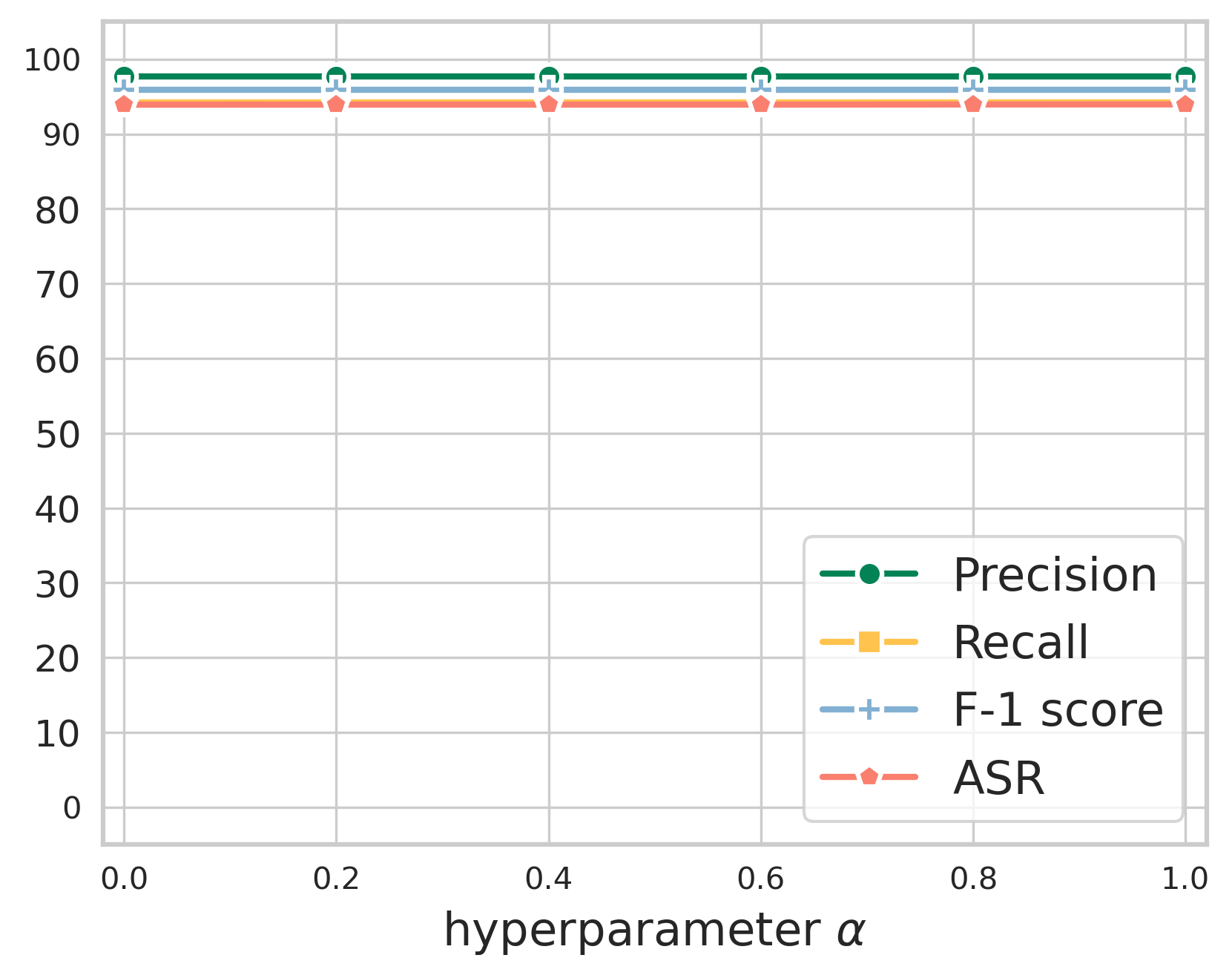}
        \caption{Deep SVDD on BGL}
        \label{fig: deepsvdd_alpha_bgl}
    \end{subfigure}
    \hfill
    \begin{subfigure}[b]{0.23\textwidth}
      \centering
        \includegraphics[width=0.95\textwidth]{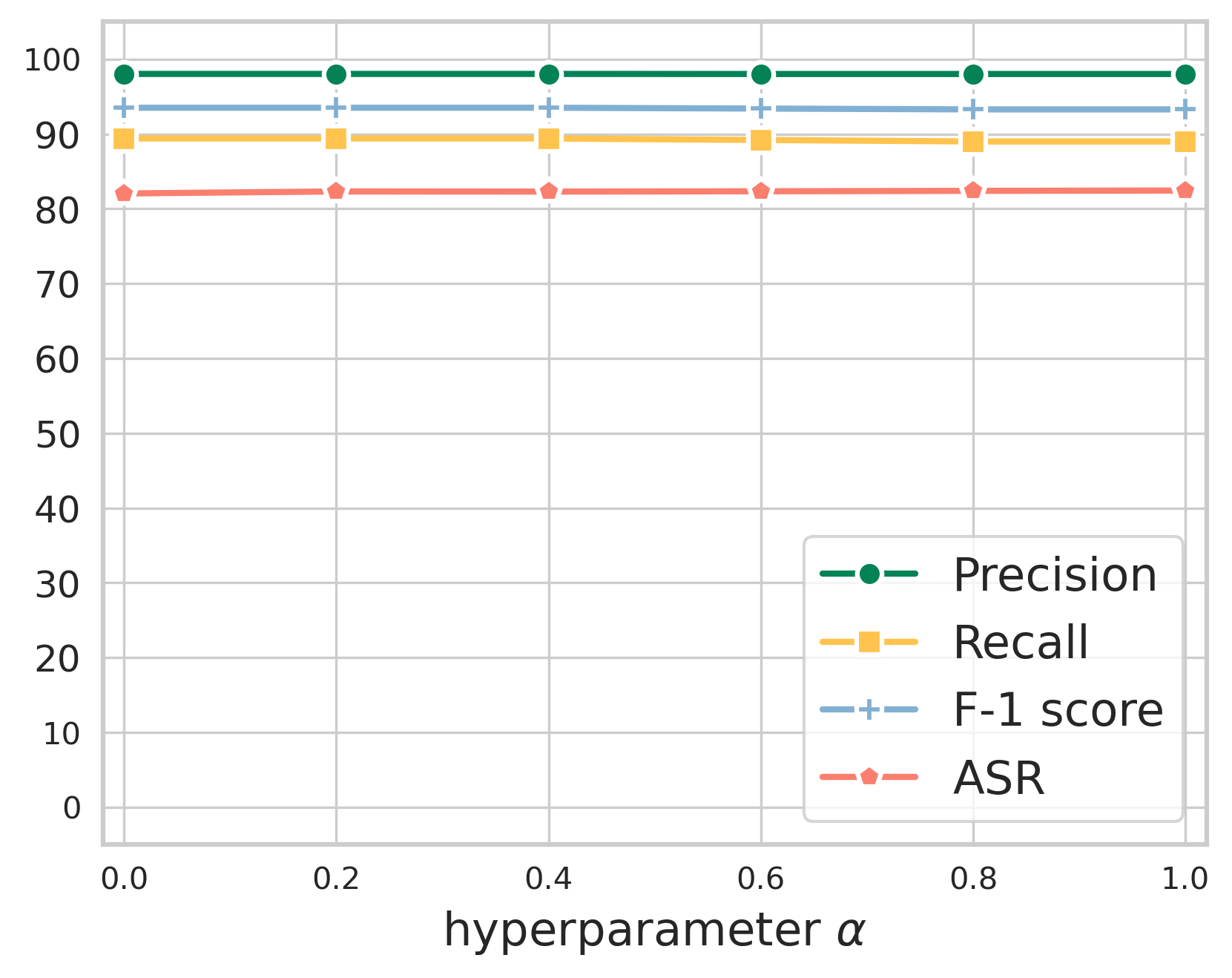}
      \caption{Deep SVDD on Thunderbird}
      \label{fig: deepsvdd_alpha_tb}
    \end{subfigure}
    \begin{subfigure}[b]{0.23\textwidth}
        \centering
        \includegraphics[width=0.95\textwidth]{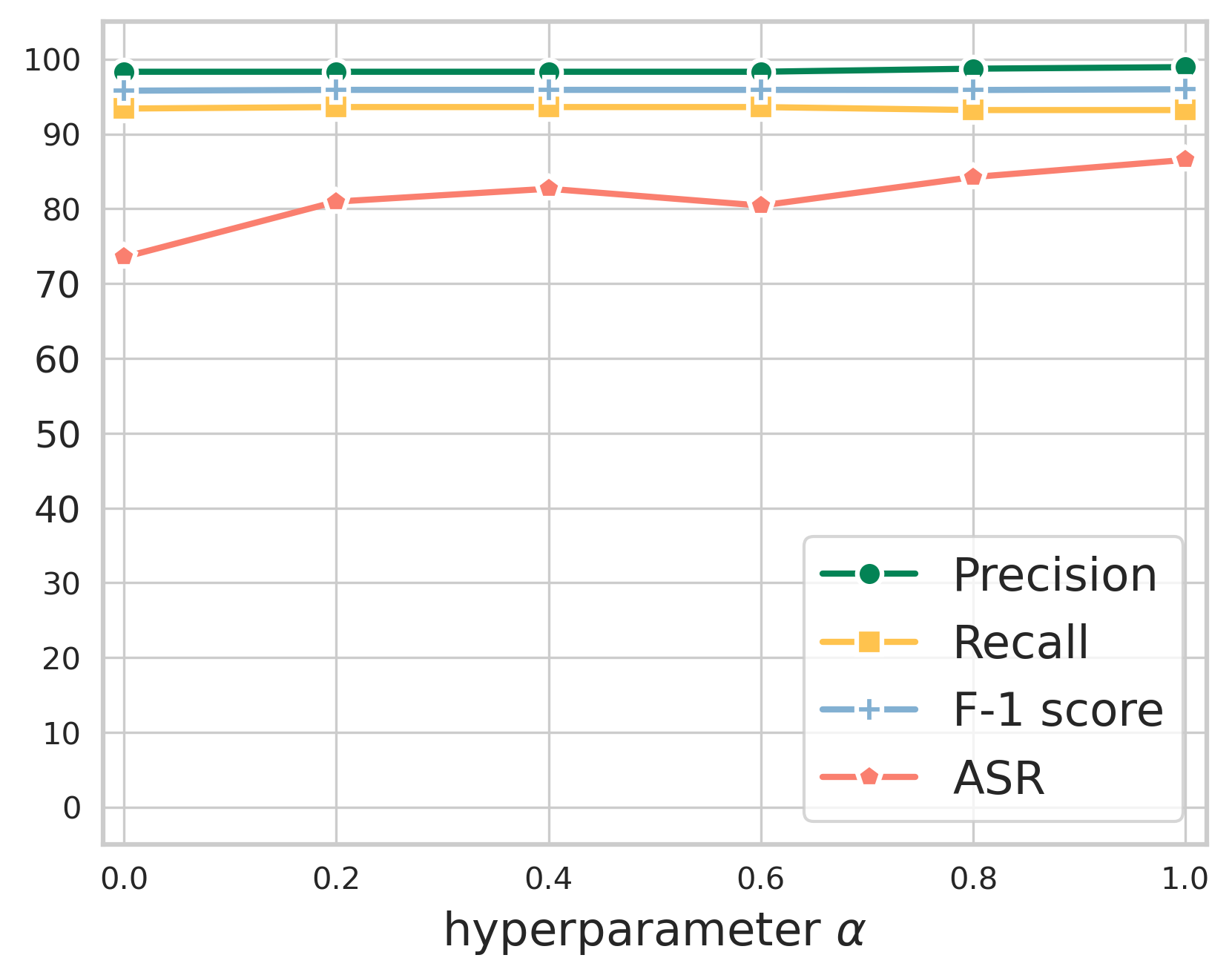}
        \caption{OC4Seq on BGL}
        \label{fig: oc4seq_alpha_bgl}
    \end{subfigure}
    \hfill
    \begin{subfigure}[b]{0.23\textwidth}
      \centering
        \includegraphics[width=0.95\textwidth]{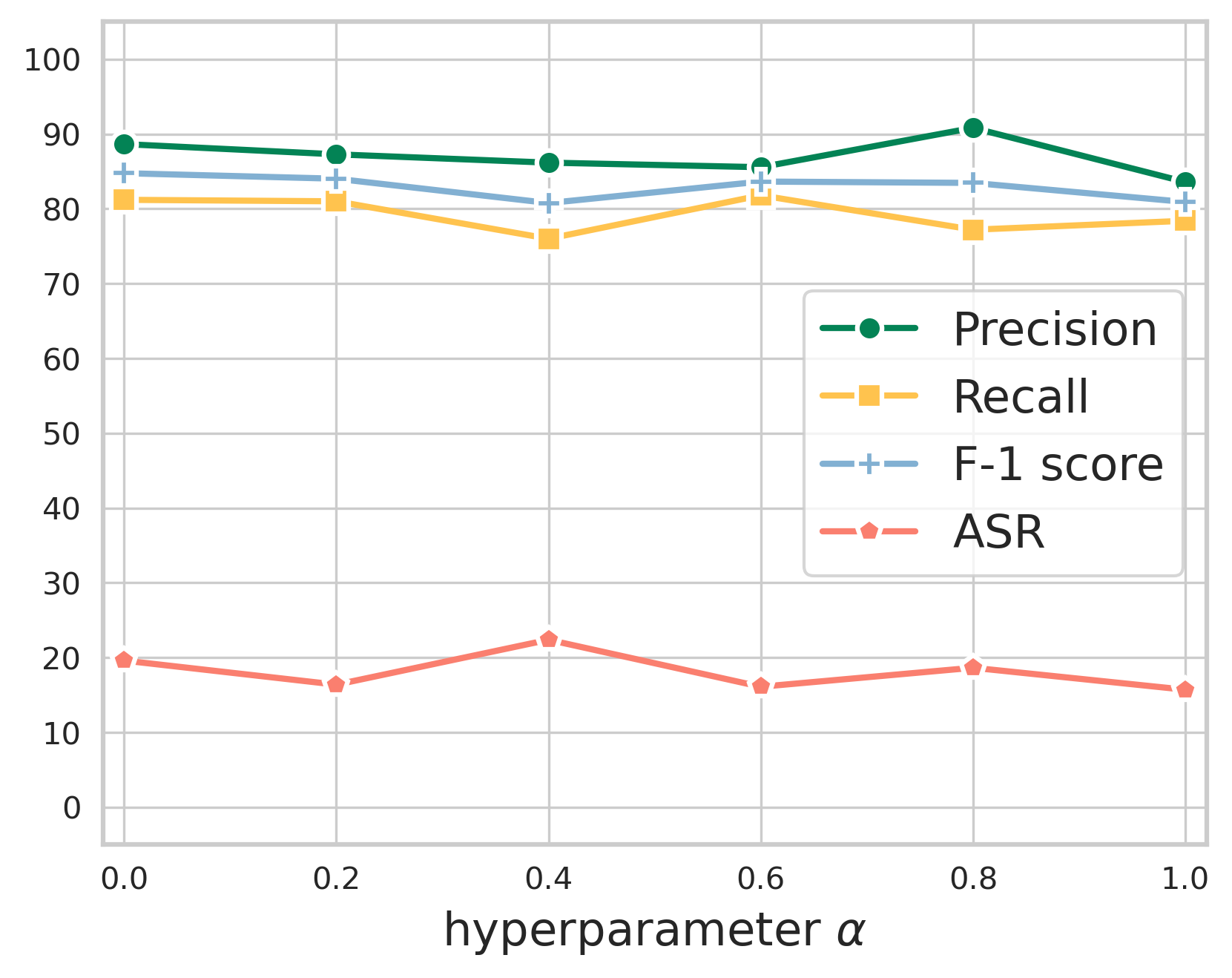}
      \caption{OC4Seq on Thunderbird}
      \label{fig: oc4seq_alpha_tb}
    \end{subfigure}  
  \caption{Results of backdoor attack for various hyperparameter $\alpha$.}
  \label{fig: sensitivity_alpha}
\end{figure}

\noindent \textit{Hyperparameter $\alpha$.}
We analyze the impact of the parameter $\alpha$ by varying its values from 0 to 1.0 in increments of 0.2. As illustrated in Figure \ref{fig: sensitivity_alpha}, increasing the values of $\alpha$ generally leads to a slight increase in ASR. For both models, ASR consistently stabilizes at a higher level with different values of $\alpha$.  It is noticeable that when $\alpha=0$, this case differs from a benign setting, as the training dataset contains perturbed samples and meanwhile, the new learning objective function still includes the perturbed sequence representation drifting term.

\begin{figure}[ht]
    \centering
    \captionsetup{justification=centering}
    \begin{subfigure}[b]{0.23\textwidth}
        \centering
        \includegraphics[width=0.95\textwidth]{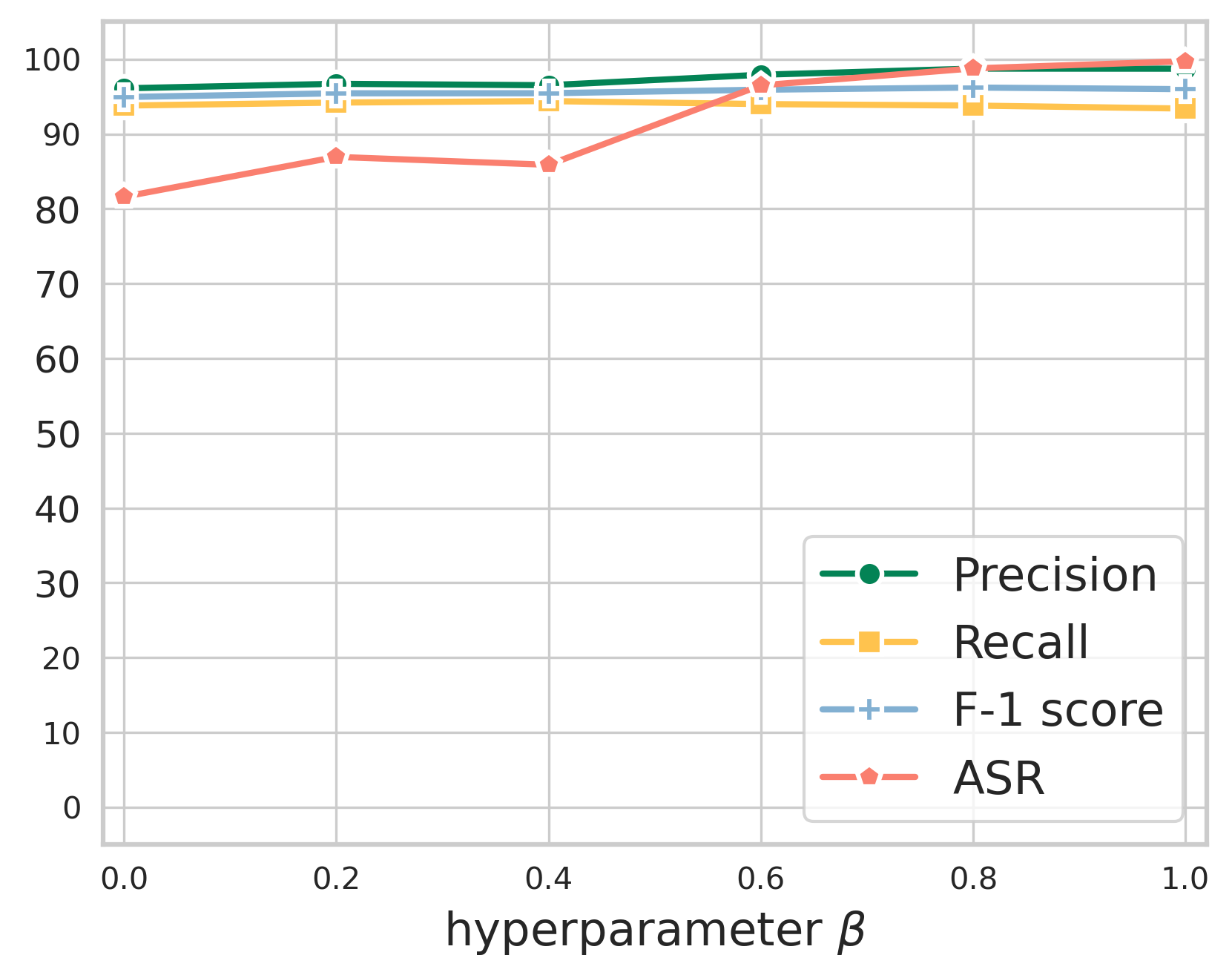}
        \caption{Deep SVDD on BGL}
        \label{fig: deepsvdd_beta_bgl}
    \end{subfigure}
    \hfill
    \begin{subfigure}[b]{0.23\textwidth}
      \centering
        \includegraphics[width=0.95\textwidth]{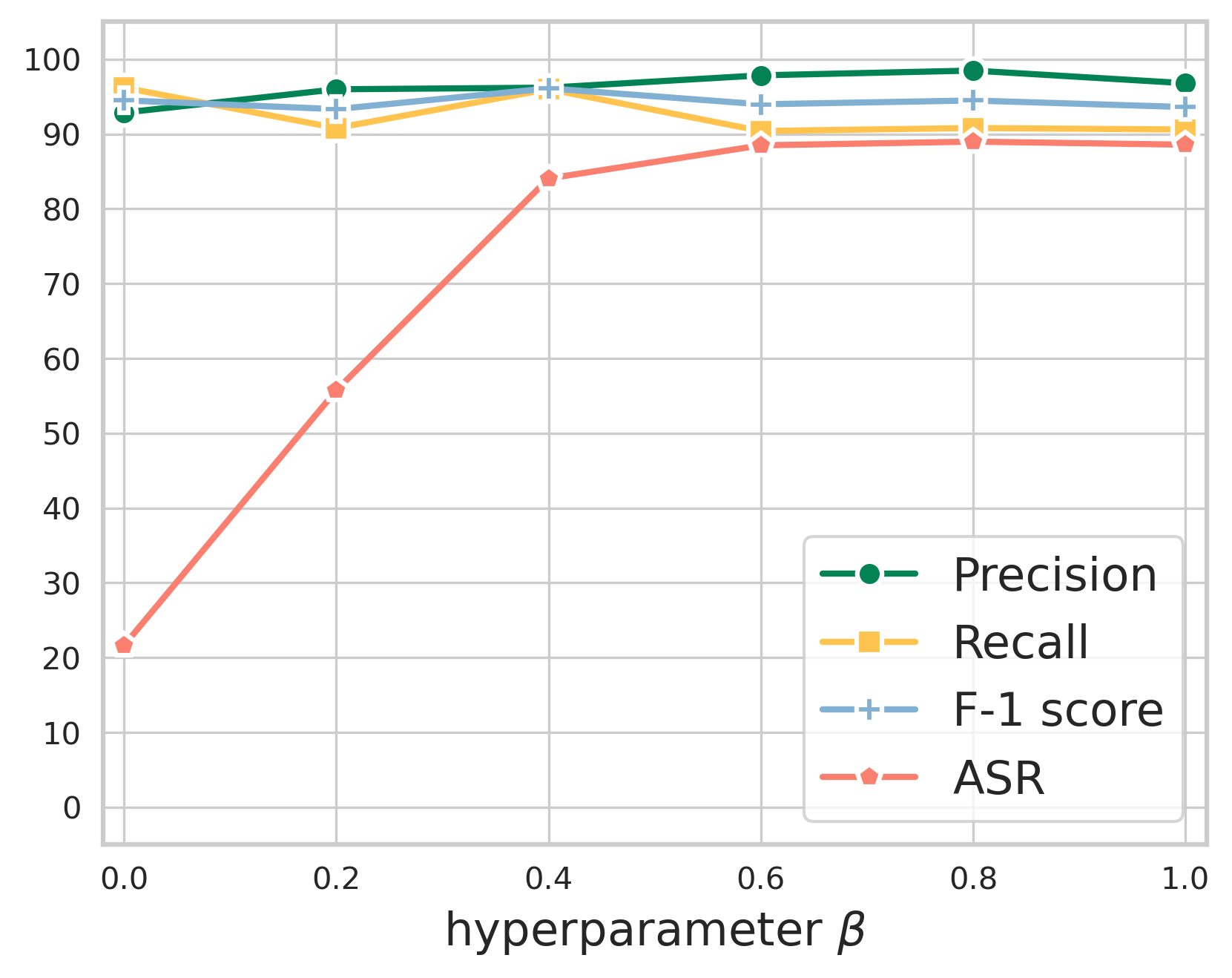}
      \caption{Deep SVDD on Thunderbird}
      \label{fig: deepsvdd_beta_tb}
    \end{subfigure}
    \begin{subfigure}[b]{0.23\textwidth}
        \centering
        \includegraphics[width=0.95\textwidth]{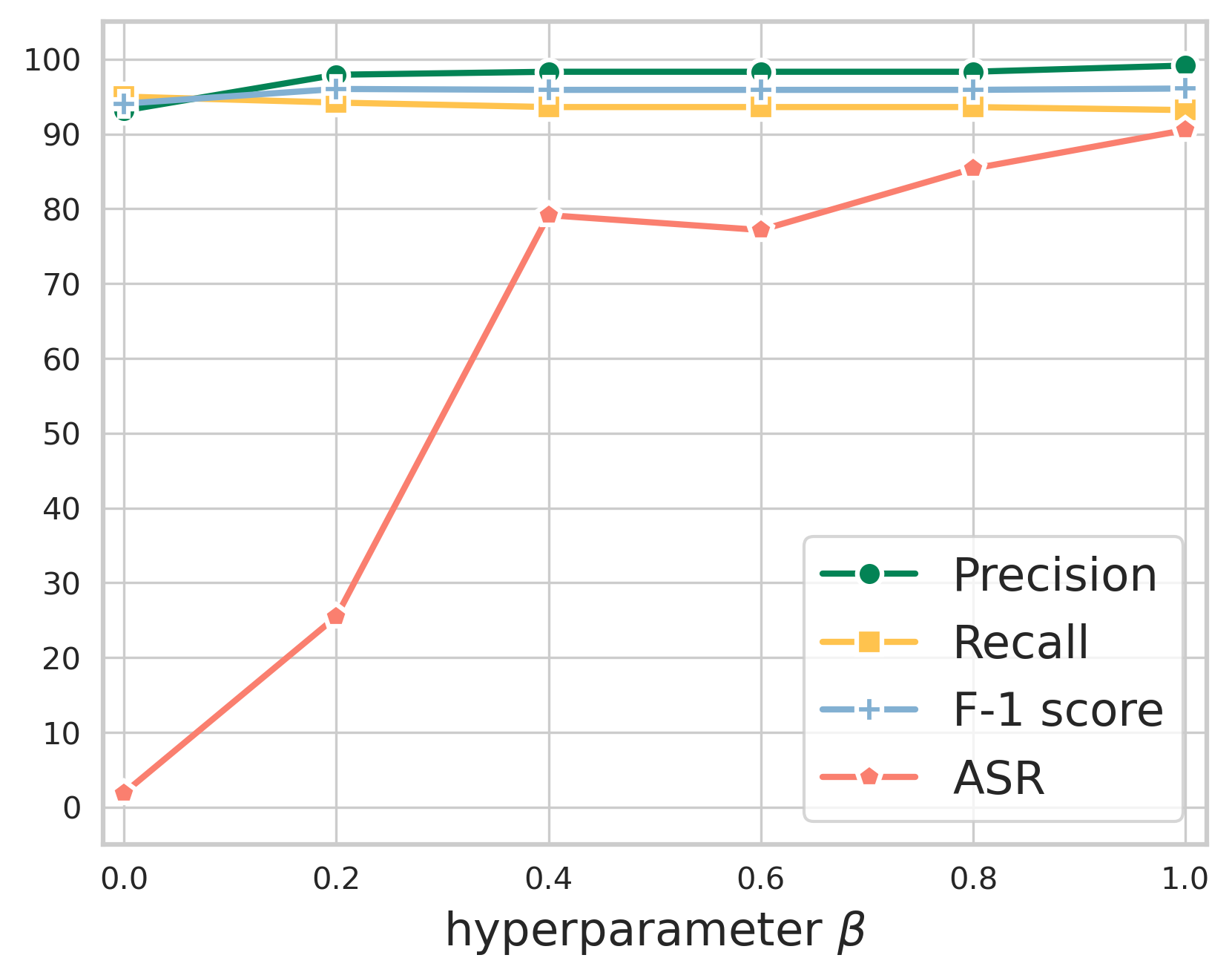}
        \caption{OC4Seq on BGL}
        \label{fig: oc4seq_beta_bgl}
    \end{subfigure}
    \hfill
    \begin{subfigure}[b]{0.23\textwidth}
      \centering
        \includegraphics[width=0.95\textwidth]{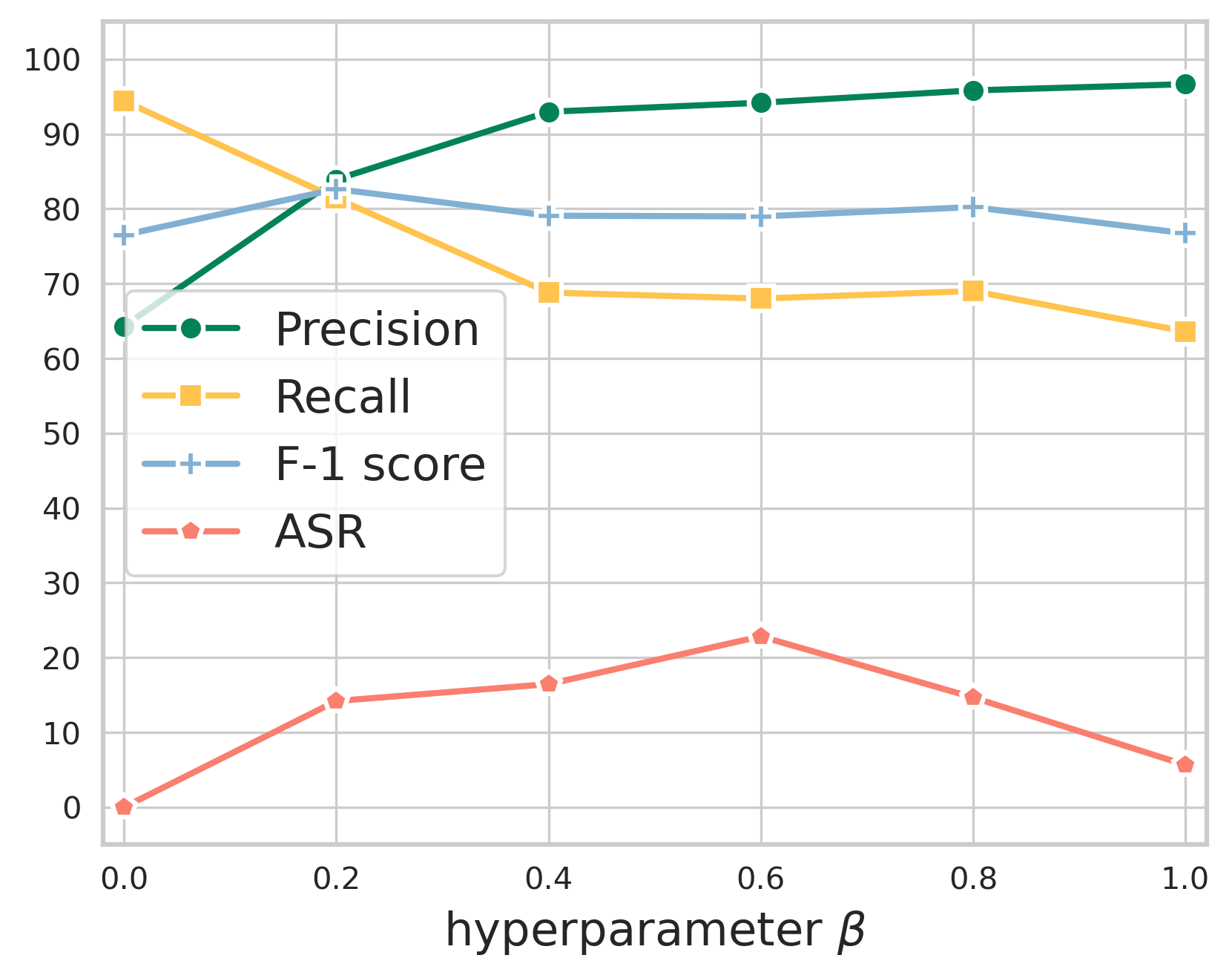}
      \caption{OC4Seq on Thunderbird}
      \label{fig: oc4seq_beta_tb}
    \end{subfigure}  
  \caption{Results of backdoor attack for various hyperparameter $\beta$.}
  \label{fig: sensitivity_beta}
\end{figure}

\noindent \textit{Hyperparameter $\beta$}.
We also investigate the impact of the parameter $\beta$ by varying its value from 0 to 1.0 in increments of 0.2. The results are presented in Figure \ref{fig: sensitivity_beta}. It is noticeable that with the increase of $\beta$, the ASR generally keeps rising and then stabilizes at a high level. For OC4Seq on Thunderbird, the ASR starts to decrease when $\beta > 0.6$.

\subsubsection{Visualization}
\begin{figure}[ht]
    \captionsetup{font={small},justification=justified}
    \centering
    \captionsetup{justification=centering}
    \begin{subfigure}[t]{0.23\textwidth}
        \centering
        \includegraphics[width=0.95\textwidth]{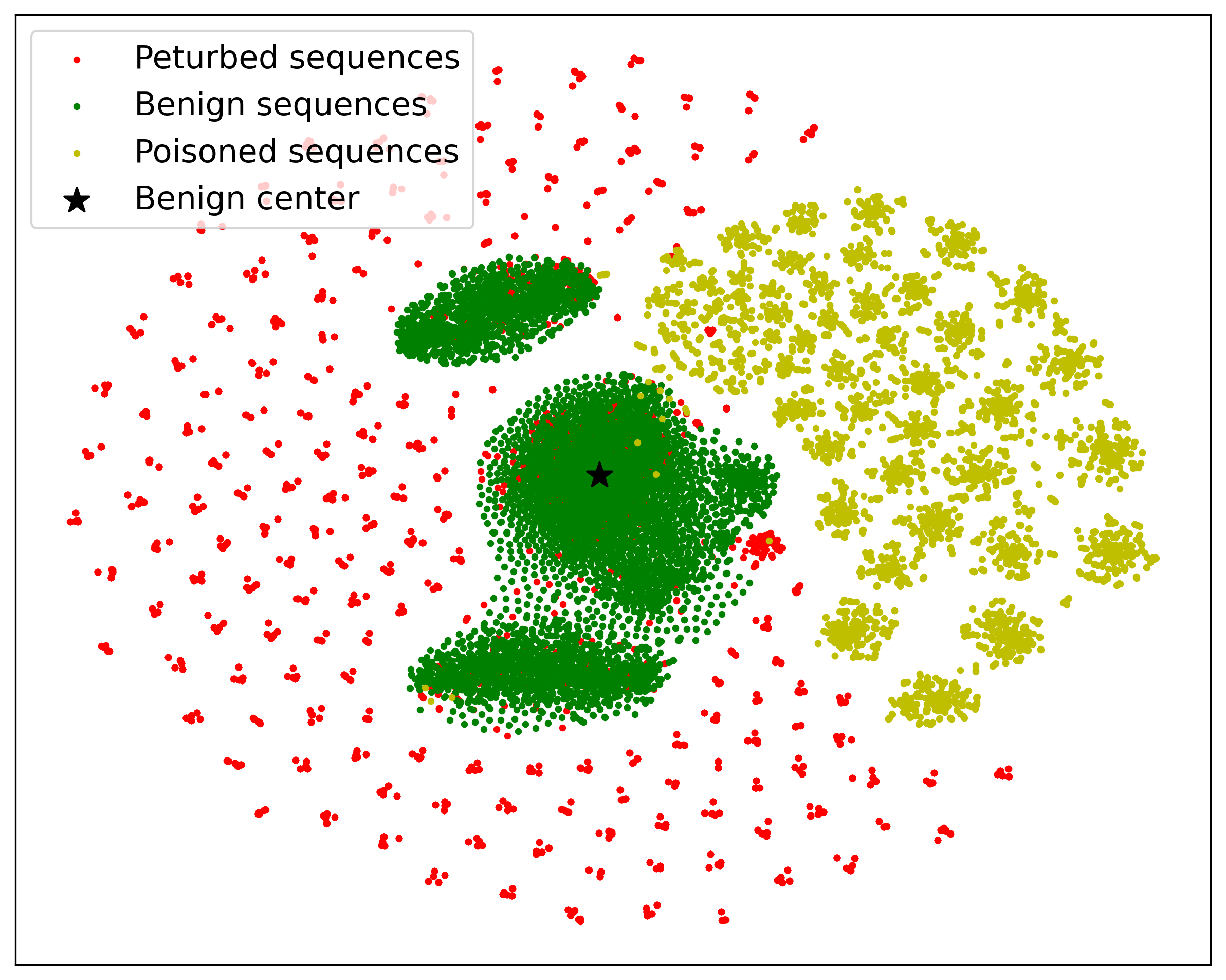}
        \caption{Benign model} 
        \label{fig: visual_benign}
    \end{subfigure}
    \begin{subfigure}[t]{0.23\textwidth}
      \centering
        \includegraphics[width=0.95\textwidth]{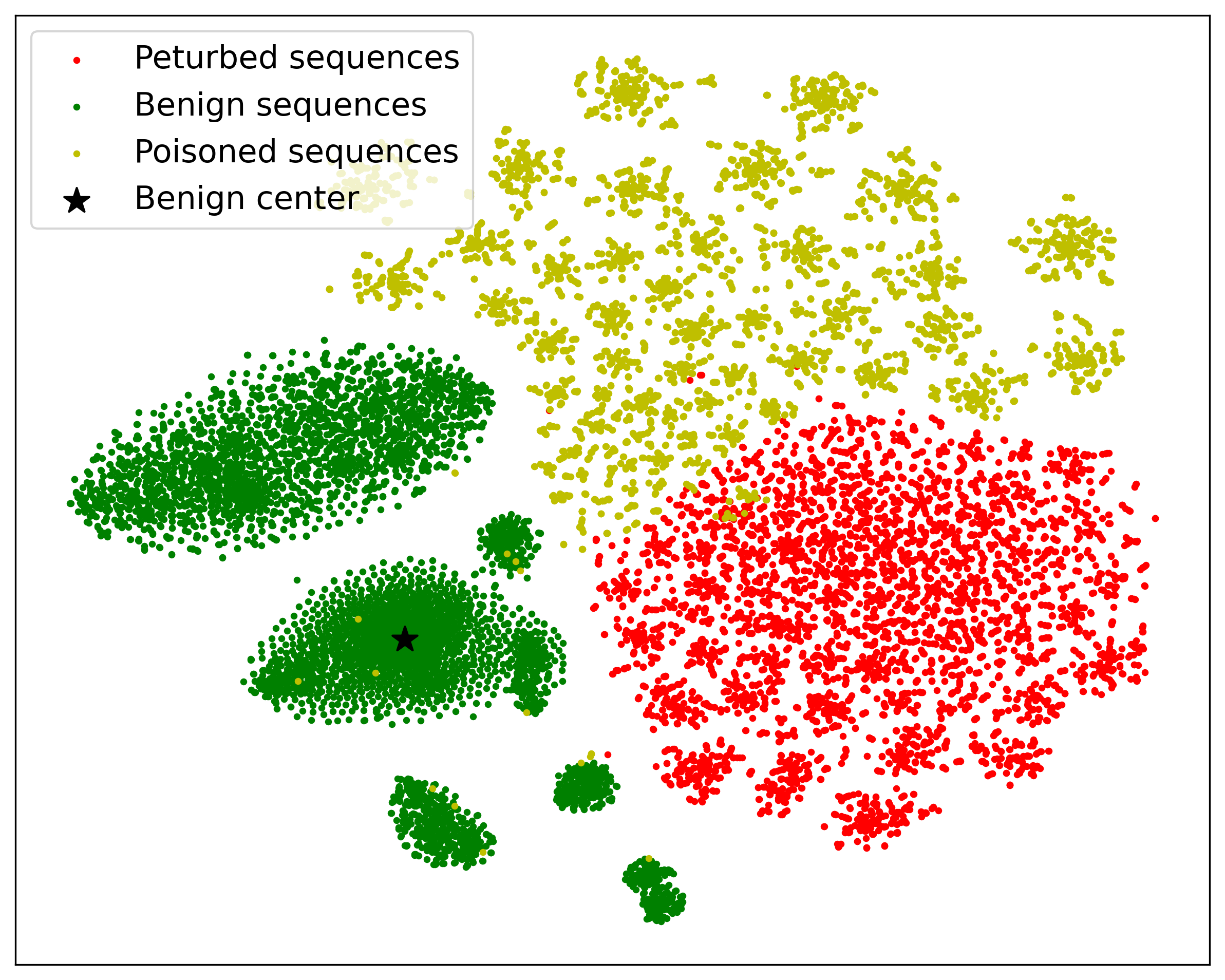}
      \caption{Infected model with $\beta = 0$}
      \label{fig: visual_c}
    \end{subfigure}
    \begin{subfigure}[t]{0.23\textwidth}
        \centering
        \includegraphics[width=0.95\textwidth]{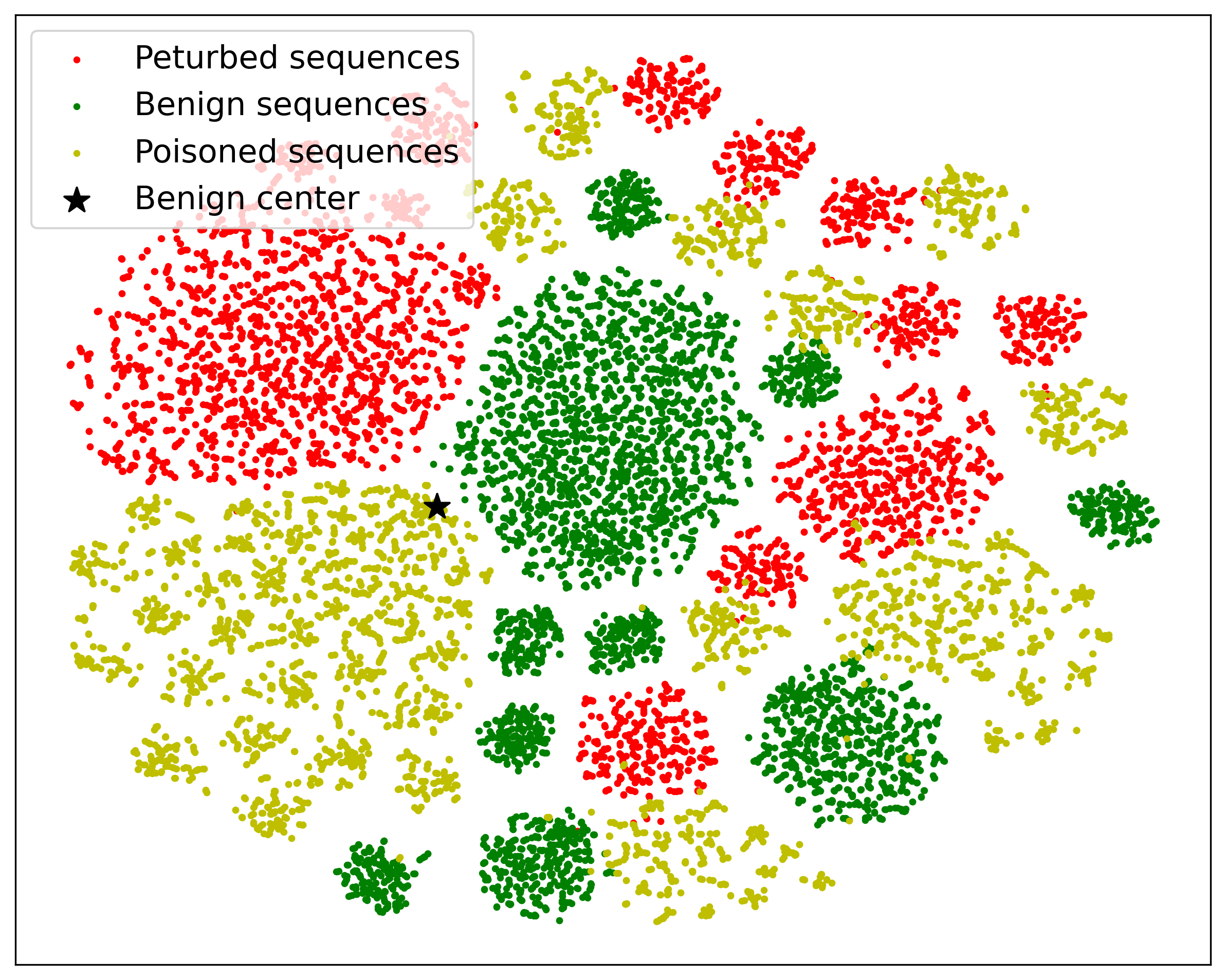}
        \caption{Infected model with $\alpha = 0$}
        \label{fig: visual_mi}
    \end{subfigure}
    \begin{subfigure}[t]{0.23\textwidth}
      \centering
        \includegraphics[width=0.95\textwidth]{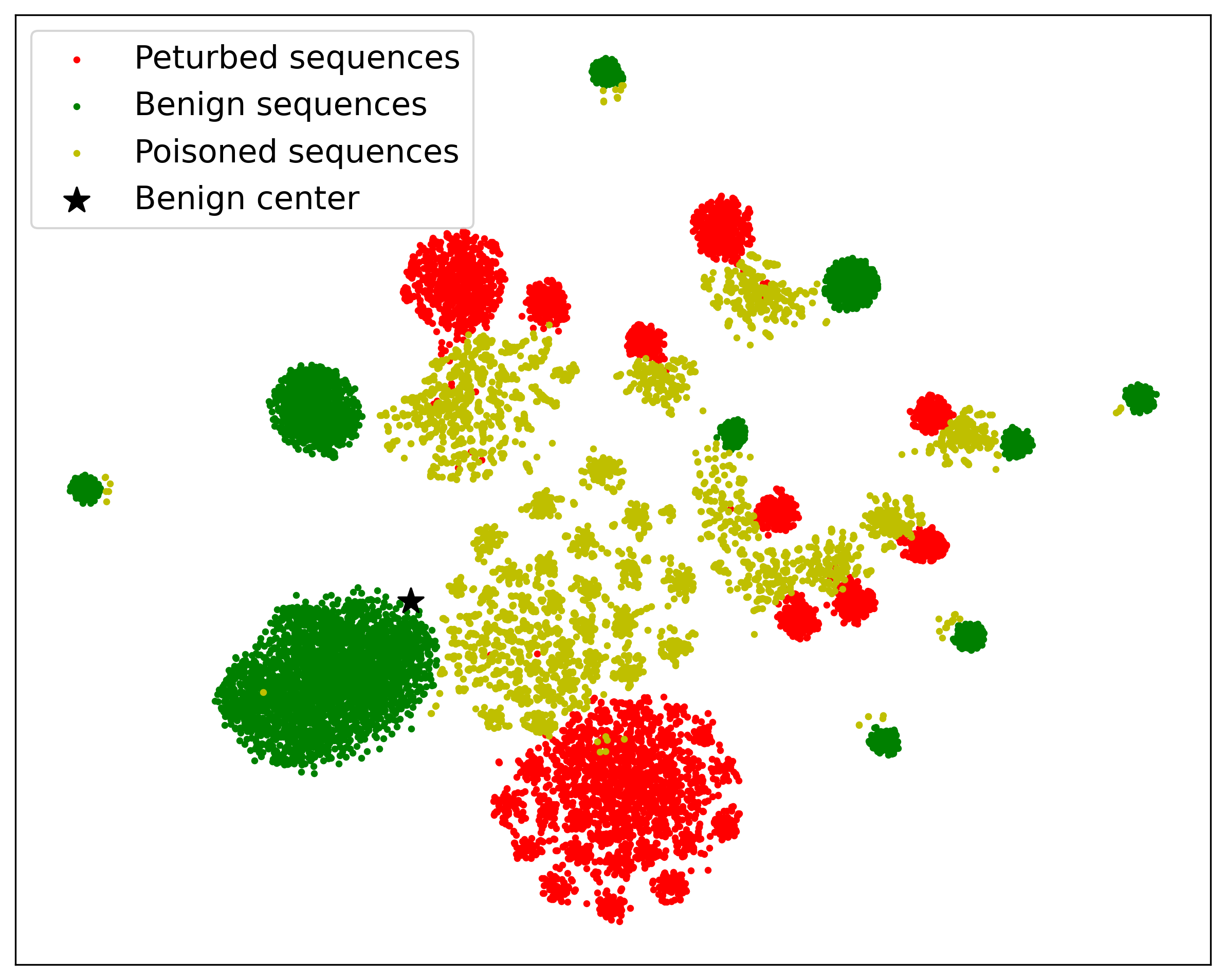}
      \caption{Infected model with $\alpha = 0.5, \beta = 0.5$}
      \label{fig: visual_both}
    \end{subfigure}  
  \caption{Visualization of benign, perturbed, and poisoned sequences.}
  \label{fig: visualization}
\end{figure}

We further visualize representations of benign, perturbed, and poisoned sequences for the infected Deep SVDD model in the BGL dataset. We randomly selected 5000 benign sequences to create their corresponding perturbed and poisoned sequences. The results are shown in Figure \ref{fig: visualization}.

Figure \ref{fig: visual_benign} shows that the representations of perturbed and poisoned sequences derived by the benign model are far from the benign center. Meanwhile, Figures \ref{fig: visual_c} and \ref{fig: visual_mi} reveal that by employing the proposed perturbed sequence center drifting or perturbed sequence representation drifting, we can either move perturbed sequences closer to the benign center or establish a correlation between perturbed and benign sequences, but cannot achieve both simultaneously.

Figure \ref{fig: visual_both} illustrates that using both training objectives, the infected model brings perturbed sequences close to the benign center and all the poisoned samples are also close to the perturbed counterpart, making them challenging to detect. Figures \ref{fig: visual_c}, \ref{fig: visual_mi}, and \ref{fig: visual_both} also demonstrate that poisoned sequences are consistently associated with their corresponding perturbed sequences, providing evidence that the models ignore the placeholders and focus on the patterns outlined in our proposed trigger generation.

\section{Conclusions}
In this paper, we have developed a novel backdoor attack framework against one-class anomaly detection models on sequential data, enabling anomalies to evade detection. Our framework comprises two essential components, trigger generation and backdoor injection. Trigger generation is to derive imperceptible backdoor triggers from the normal sequences, while the backdoor injection is to inject backdoor patterns to infected models during the training phase by developing two learning objectives. After deployment, the attacker can conceal abnormal entries within the sequences, which enables anomalies to evade detection by the infected model. Our experiments on one-class anomaly detection models demonstrate the effectiveness of our proposed backdoor attack strategy. In the future, we plan to study how to effectively defend the backdoor attack against the sequential anomaly detection models.

\section*{Acknowledgement}
This work was supported in part by NSF 2103829.

\end{document}